\documentclass[conference]{IEEEtran}
\IEEEoverridecommandlockouts
\usepackage{cite}
\usepackage{amsmath,amssymb,amsfonts}

\usepackage{algorithm}
\usepackage{algpseudocode} 
\usepackage{graphicx}
\usepackage{subcaption}
\usepackage{soul}
\usepackage[acronym]{glossaries}
\graphicspath{ {./Images/} }
\usepackage{textcomp}
\usepackage{xcolor}
\usepackage{comment}
\def\BibTeX{{\rm B\kern-.05em{\sc i\kern-.025em b}\kern-.08em
    T\kern-.1667em\lower.7ex\hbox{E}\kern-.125emX}}
\DeclareMathOperator*{\argmax}{arg\,max}

\makeglossaries
\newacronym{SL}{SL}{sidelink}
\newacronym{V2V}{V2V}{vehicle to vehicle}
\newacronym{QoS}{QoS}{quality of service}
\newacronym{V2I}{V2I}{vehicle to infrastructure}
\newacronym{V2X}{V2X}{vehicle to everything}
\newacronym{DQN}{DQN}{deep \textit{Q}-network}
\newacronym{DDQN}{DDQN}{double-DQN}
\newacronym{DNN}{DNN}{deep neural network}
\newacronym{TQL}{TQL}{transfer \textit{Q}-learning}
\newacronym{GEMV}{GEMV$^{2}$}{Geometry-based Efficient propagation model for V2V communication}
\newacronym{CSI}{CSI}{channel state information}
\newacronym{RL}{RL}{reinforcement learning}
\newacronym{TL}{TL}{transfer learning}
\newacronym{MARL}{MARL}{multi-agent reinforcement learning}
\newacronym{OSM}{OSM}{OpenStreetMap}
\newacronym{SUMO}{SUMO}{Simulation of Urban MObility}
\newacronym{SINR}{SINR}{signal to interference plus noise ratio}
\newacronym{MDP}{MDP}{Markov decision process}
\newacronym{BS}{BS}{base station}
\newacronym{DQL}{DQL}{deep \textit{Q}-learning}
\newacronym{GCM}{GCM}{geometric channel models}
\newacronym{D2D}{D2D}{device-to-device}
\newacronym{ReLU}{ReLU}{rectified linear unit}
 
\begin{document}

\title{Transfer Learning in Multi-Agent Reinforcement Learning with Double Q-Networks for Distributed Resource Sharing in V2X Communication\\}

\author{\IEEEauthorblockN{Hammad Zafar, Zoran Utkovski, Martin Kasparick, S{\l}awomir Sta\'nczak}
\IEEEauthorblockA{\textit{Wireless Communications and Networks, 
Fraunhofer Heinrich Hertz Institute Berlin, Germany} \\}

}


\maketitle

\begin{abstract}
This paper addresses the problem of decentralized spectrum sharing in  vehicle-to-everything (V2X) communication networks. The aim is to provide resource-efficient coexistence of vehicle-to-infrastructure (V2I) and vehicle-to-vehicle (V2V) links. A recent work on the topic proposes a multi-agent reinforcement learning (MARL) approach based on deep Q-learning, which leverages a fingerprint-based deep Q-network (DQN) architecture. This work considers an extension of this framework by combining Double Q-learning (via Double DQN) and transfer learning. The motivation behind is that Double Q-learning can alleviate the problem of overestimation of the action values present in conventional Q-learning, while transfer learning can leverage knowledge acquired by an expert model to accelerate learning in the MARL setting. The proposed algorithm is evaluated in a realistic V2X setting, with synthetic data generated based on a geometry-based propagation model that incorporates  location-specific geographical  descriptors  of the simulated environment (outlines of buildings, foliage, and vehicles). The advantages of the proposed approach are demonstrated via numerical simulations.

\end{abstract}

\begin{IEEEkeywords}
V2X communication, spectrum sharing, multi-agent reinforcement learning, deep Q-learning, transfer learning.
\end{IEEEkeywords}

\section{Introduction}
In vehicle-to-everything (V2X) networks,  vehicle-to-infrastructure
(V2I) and vehicle-to-vehicle (V2V) connections are supported through cellular and sidelink radio interfaces respectively.  With the introduction of use cases such as tele-operated driving, in-vehicle entertainment and automated driving, \acrshort{V2X} networks are required to provide simultaneous
support for mobile high-data rates and advanced
driving, i.e.~safety-related services. For example, while the in-vehicle entertainment applications
require high bandwidth V2I connection to the base station (BS), safety applications need to periodically disseminate messages among neighboring vehicles with high reliability and low latency. 

To provide resource-efficient coexistence
of V2I and V2V connections, sidelink V2V connections may be configured to share spectrum
with V2I links. In that case, effective strategies for spectrum sharing with
V2I links are required, including the selection of spectrum sub-band and control of transmission power, in order to meet the diverse
service requirements of both V2I and V2V links. Furthermore, taking into account the nature of the \acrshort{V2V} traffic, decentralized solutions are favored over centralized schemes that come with large transmission overhead and higher latency. 

The decentralized spectrum sharing problem can  be cast as a multi-agent reinforcement
learning (MARL) problem where the V2V links, each acting as an agent, collectively interact with the communication environment and learn to improve
spectrum and power allocation by 
using the gained experiences. Along these lines, a MARL approach based 
on deep Q-learning has been proposed in~\cite{IEEEexample:liang2019spectrum}. The approach leverages a fingerprint-based deep Q-network (DQN) 
method that is amenable to a distributed implementation, thus offering a potential solution to the decentralized spectrum sharing problem. 



Reinforcement learning methods based on Q-learning/deep Q-learning are known to sometimes learn unrealistically high action values, which might have a negative effect on policy quality and thus on the overall performance in the studied V2X scenario. In addition, RL algorithms based on Q-learning typically require  a large  number  of  training  episodes  for  an  agent  to  learn  from the  environment. Moreover, in the deep Q-learning framework the  parameter  set  for  the  trained DQNs needs to be updated when the environment conditions significantly change. Hence, decreasing the complexity/speeding-up the training process is of practical interest in vehicular scenarios with varying channel conditions and network topologies.

To address these issues, we propose an extension of the MARL framework in  \cite{IEEEexample:liang2019spectrum}, by combining the principles of Double Q-learning (Double DQN  \cite{IEEEexample:van2016deep}) and transfer learning. The rationale is that Double Q-learning can prevent overestimation of the action values, while transfer learning can leverage knowledge acquired by an expert model to reduce the duration of the training phase of the learner model. In the proposed framework, the expert model learns its policies using the \acrshort{DDQN} algorithm. The learner model, which also employs \acrshort{DDQN}, uses the Q-values from the expert model to adjust its network parameters in the direction that improves the system performance. 

To evaluate the performance of the proposed framework in a realistic vehicular setting, we will resort to the 
$\mathrm{GEMV}^2$ model (short for Geometry-based Efficient propagation Model for V2V communication) \cite{IEEEexample:boban2014geometry}. In contrast to stochastic models such as, e.g., the urban channel model from Annex A in 3GPP TR 36.885 used in \cite{IEEEexample:liang2019spectrum}, geometry-based deterministic  models  
intrinsically  support  time  evolution  and  spatial consistency, making them appropriate for evaluation of MARL algorithms.

\section{System Model}
\label{System Model}
 We adopt the system model from \cite{IEEEexample:liang2019spectrum}, where we consider a set of $M$ V2I links, denoted by $\mathcal{M} =\left \{ 1,2,...,M \right \}$, and a set of $K$ \acrshort{V2V} links, denoted by $\mathcal{K} =\left \{ 1,2,...,K \right \}$, in a cellular network with a single \acrfull{BS}. Each of the $K$ \acrshort{V2V} links correspond to a pair of communicating \acrshort{V2V} users.
This model is synonymous with 3GPP \acrshort{V2X} architectures \cite{IEEEexample:Paper_12}, where \acrshort{V2I} and \acrshort{V2V} connections are supported using cellular and NR sidelink radio interfaces, respectively.

We assume that the $M$ \acrshort{V2I} links have preassigned orthogonal spectrum sub-bands (a sub-band being a block of consecutive sub-carriers) and fixed transmission power, while each of the $K$ \acrshort{V2V} links can only occupy a single spectrum sub-band. The resource allocation task for the $K$ \acrshort{V2V} links is to learn efficient policies to choose transmission powers and spectrum sub-bands. During one coherence time period of length $\Delta_T$, the channel
power gain, $g_k[m]$, of the $k$-th V2V link over the $m$-th sub-band (occupied by the $m$-th V2I link) follows $g_k[m] = \alpha_k h_k[m]$, where $\alpha_k$ captures the large-scale signal attenuation (large-scale fading) and $h_k[m]$ is the small-scale fading
power component.  The small-scale fading power component is modeled as an exponentially distributed random variable with unit mean, whose realization is assumed to remain constant during the coherence time period. 


With the above, the interference to the $m$-th \acrshort{V2I} link is caused by the \acrshort{V2V} links using the same sub-band. Following the notation in~\cite{IEEEexample:liang2019spectrum}, the received \acrfull{SINR} for the $m$-th \acrshort{V2I} link is given as

\begin{equation}
\gamma_{m}^{c} \left [ m \right ] = \frac{P_{m}^{c}\hat{g}_{m,B}[m]}{\sigma ^{2} +  \sum\limits_{k} \rho _{k}[m]P_{k}^{d}[m]g_{k,B}[m]},
\label{eq:SINR_V2I}
\end{equation}
where, $P_{m}^{c}$ and $P_{k}^{d}[m]$ denotes the transmit power for the $m$-th \acrshort{V2I} transmitter and the $k$-th \acrshort{V2V} link transmitter over the $m$-th sub-band, respectively. We use $\hat{g}_{m,B}$ to denote the channel gain of the $m$-th V2I link and $g_{k,B}$ to denote the gain of the interfering channel from the $k$-th V2V transmitter
to the BS over the $m$-th sub-band. $\sigma ^{2}$ is the noise power and $\rho_{k}[m]$ is the spectrum allocation indicator with $\rho_{k}[m]=1$ indicating that the $k$-th \acrshort{V2V} link uses the $m$-th sub-band, and $\rho_{k}[m]=0$ otherwise.     

Similarly, the received \acrshort{SINR} of the $k$-th V2V
link over the $m$-th sub-band is expressed as 
\begin{equation}
\gamma _{k}^{d} \left [ m \right ] = \frac{P_{k}^{d}[m]g_{k}[m]}{\sigma ^{2} + I_{k}[m]},
\end{equation}
where, $g_{k}[m]$ is the channel between $k$-th \acrshort{V2V} link transmitter and receiver, $I_{k}[m]$ denotes the interference power experienced by the $k$-th V2V link receiver on the $m$-th sub-band, given as 


 \begin{equation}
 I_{k}[m]= {P_{m}^{c}\hat{g}_{m,k}[m]} + \sum_{k'\neq  k} \rho _{k'}[m]P_{k'}^{d}[m]g_{k',k}[m],
 \label{eq:interference_power}
\end{equation}
In (\ref{eq:interference_power}), we use $\hat{g}_{m,k}[m]$ to denote the interfering channel from the $m$-th V2I transmitter to the $k$-th V2V receiver over the $m$-th sub-band, and $g_{k',k}$ to denote the interfering channel from the $k'$-th V2V transmitter to the $k$-th V2V receiver over the $m$-th sub-band. We assume
each V2V link only accesses one sub-band, i.e., $\sum_{m=1}^M \rho_k[m]\leq 1$.

\section{Multi-agent-reinforcement learning for spectrum sharing in \acrshort{V2X}}
\label{DRL}


A multi-agent reinforcement learning (MARL) approach based on deep \textit{Q}-learning was proposed in~\cite{IEEEexample:liang2019spectrum} as a solution to the distributed spectrum sharing problem in \acrshort{V2X} communications. As our contribution focuses on an extension of this framework, in the following we summarize its main aspects.


In the MARL setting, each \acrshort{V2V} link acts as an independent agent and interacts with the environment to gain experience,  
which is then used for its own policy design. The resource sharing problem is modeled as a cooperative game through using the global reward
for all agents in the interest of an improved global network performance. For the V2I links, the objective is to maximize the sum-rate  $\sum_{m} C_m^c[m]$, where 
\begin{equation}
    C_m^c[m]=W\log(1+\gamma_m^c[m])
    \label{eq:sum-rate}
\end{equation}
is the rate that can be supported on the $m$-th V2I link with bandwidth $W$ and SINR $\gamma_m^c[m]$ as given by (\ref{eq:SINR_V2I}). For the V2V links, on the other hand, of interest is the V2V payload delivery rate, i.e.  the success probability of delivering packets of a certain size within
a certain time budget. Considering a block-fading scenario with channel coherence time $\Delta_T$, packet size $B$ (in bits) and time budget $T$ expressed in multiples of the coherence time $\Delta_T$, the V2V payload delivery rate is given by 
\begin{equation}
    \mathbb{P}\left[\sum_{t=1}^T\sum_{m=1}^M \rho_k[m] C_k^d[m,t]\geq \frac{B}{\Delta_T} \right],\:k\in\mathcal{K},
\end{equation}
where we have defined 
\begin{equation}
    C_k^d[m,t]=W\log(1+\gamma_k^d[m,t])
\end{equation}
to be the rate that can be supported on the $k$-the V2V link (operating over the $m$-th sub-band) with SINR $\gamma_k^d[m,t]$, within the $t$-th coherence time interval $\Delta_T$.

The environment is modeled as a \acrfull{MDP}\cite{IEEEexample:watkins1992q}. At each time-step $t$, each \acrshort{V2V} agent $k\in\mathcal{K}$ captures local observation $S_t^{(k)}$ of the environment and takes an action $A_{t}^{(k)}$ (comprising the selection of sub-band and transmission power), contributing to a joint action $\mathbf{A}_t$. Thereafter, each agent receives a global reward $R_{t+1}$ and the  environment changes to the next state $S_{t+1}$. 
The local observation of the environment captured by the $k$-th \acrshort{V2V} agent is expressed as  
 \begin{equation}
 \label{eqn:Modified_State_Eq}
S_{t}^{(k)}= \left \{ \{G_k[m]\}_{m\in\mathcal{M}}, \{I_{k}[m]\}_{m\in\mathcal{M}}, B_{k}, T_{k},\epsilon,i\right\},
\end{equation}
where, $G_k[m]$ captures local channel information 
\begin{equation*}
G_k[m]=\{g_{k,B}[m], \hat{g}_{m,k}[m], g_{k',k}[m], g_{k}[m]\},
\end{equation*}
 $B_{k}$ is the remaining payload, and $T_{k}$ is the remaining time budget (in multiples of $\Delta_T$). The $\epsilon$ of $\epsilon$-greedy strategy \cite{IEEEexample:sutton2018reinforcement} and $i$, iteration number of training are used as a low-dimensional fingerprint to track the policy changes. \smallskip


\noindent \textbf{Reward design:} 
At each time step $t$, the global reward function is designed as a combination of a V2I-related reward that equals the sum-rate $C_m^c[m]$ in (\ref{eq:sum-rate}), and a V2V-related reward expressed as


\begin{equation}
L_t^{(k)}= \begin{cases}
       \sum\limits_{m=1}^{M} \rho _{k}[m] C_k^d[m,t], & \text{if}\ B_{k}\geq 0, \\
      \beta, & \text{otherwise}.
    \end{cases}
    \label{eq:V2V_award}
\end{equation}
In effect, the V2V-related reward is set to be equal to the effective V2V transmission rate until the payload
is delivered, after which the reward is set to a constant number, $\beta$, that is greater than the largest possible V2V transmission
rate. In practice, $\beta$ is a hyper-parameter that needs to be tuned
empirically.

With this, the global reward at each time step reads
\begin{equation}
R_{t+1} = \lambda_{c}\sum\limits_{m} C_{m}^{c}[m,t] + \lambda_{d}\sum\limits_{k} L_t^{(k)}, 
\end{equation}
where, $\lambda_c$ and $\lambda_d$ are positive weights to balance the V2I and V2V
objectives.\smallskip




\noindent \textbf{Learning procedure:} The approach in \cite{IEEEexample:liang2019spectrum} leverages deep Q-learning (DQL) with
experience replay \cite{IEEEexample:mnih2015human} and 
consists of two phases. In the learning (training) phase, the global reward is made  accessible
to each V2V agent, which then adjusts its actions
towards an optimal policy through updating its deep Q-network
(DQN). In the implementation phase, each V2V agent receives
local observations of the environment and then selects an
action according to its trained DQN. 

Each V2V agent $k$ has a dedicated DQN that takes as input
its local current observation $S_t^{(k)}$ of the environment and outputs the value functions
corresponding to all actions. 
Following the environment
transition, each agent $k$ collects and stores the transition
tuple, $\left(S_{t}^{(k)}, A_{t}^{(k)}, R_{t+1}, S_{t+1}^{(k)} \right )$ in a replay memory. 
At each episode, a mini-batch $\mathcal{B}$ of experiences is randomly sampled from the memory to update the parameter set $\boldsymbol{\theta}$ of the DQN. 
For \acrshort{DQL}, besides the (main) DQN, a target DQN is created with parameters $\boldsymbol{{\grave{\theta}}}$ that are
duplicated from the DQN parameters $\boldsymbol{\theta}$ periodically
and are kept fixed for $\tau$ updates. 
The DQN is updated using stochastic gradient descent by minimizing the loss function
\begin{equation}
\label{eq:L_DQN}
\begin{aligned}
\mathcal{L}_{\mathrm{DQN}}^{(k)}
=\sum_{\mathcal{B}^{(k)}} \, [R_{t+1} +  \gamma \,\underset{a \in \mathcal{A}^{(k)}} {\max} \, Q(S_{t+1}^{(k)},a;\boldsymbol{{\grave{\theta}}}_{t}^{(k)}) \\
    -Q(S_{t}^{(k)},A_{t}^{(k)};\boldsymbol{\theta}_{t}^{(k)})]^{2},
\end{aligned}
\end{equation}
where, 
$\gamma \in [0,1]$ is a discount factor  that trades off the importance of immediate and later rewards. 


\section{Multi-Agent Reinforcement Learning with Double DQN and Transfer Learning}
\label{Contribution}
Both Q-learning and deep Q-learning with DQN are known to overestimate action values under certain conditions. As noted in \cite{IEEEexample:hasselt2010double}, this is mostly due to the $\max$ operator in standard Q-learning and DQN, which uses the same values both to select and to evaluate an action, potentially resulting in overoptimistic value estimates. Motivated by this observation, we propose to substitute the DQN architecture in \cite{IEEEexample:liang2019spectrum} with the Double DQN architecture from~\cite{IEEEexample:van2016deep}. The expectation is that Double DQN can prevent overestimation by decoupling the selection step  from the evaluation step, potentially improving the overall performance in the V2X spectrum sharing setting.   

In addition to the aspect of overestimation of action values, in this paper, we also address solutions that aim at speeding up the learning process, as conventional RL algorithms based on Q-learning typically require  a large  number  of  training  episodes  for  an  agent  to  learn  from the  environment.  Moreover, in the DQL framework the  parameter  set  for  the  trained DQNs needs to be updated when the environmental conditions significantly change. This extensive training procedure may not be a feasible  solution  over  different  channel  conditions  and network  topologies.  To  alleviate  this  problem, we propose to incorporate transfer  learning  (TL) that harnesses previously acquired knowledge to reduce the duration of the learning phase of the \acrshort{DQN}/Double DQN algorithm. 

\subsection{Double DQN}
\label{DDQN}



In Double DQN, referred to as \acrshort{DDQN} in the following, the task of evaluation and selection of the action is
decoupled by using a double estimator approach having two parameter sets. The overestimation is reduced by decoupling the evaluation process of the Q-values from the selection of action. Although not fully decoupled, the target network in the DQN architecture provides a natural candidate for the evaluation function, without having to introduce additional networks. With this modification, the loss function for \acrshort{DDQN}, which is based on the loss function in (\ref{eq:L_DQN}), is expressed as

\begin{equation}
\begin{aligned}
\mathcal{L}^{(k)}_{\mathrm{DDQN}} 
= \sum_{\mathcal{B}^{(k)}} \, &[R_{t+1} -Q(S_{t}^{(k)},A_{t}^{(k)};\boldsymbol{\theta}_{t}^{(k)}) \\
     & +  \gamma \, Q(S_{t+1}^{(k)},\underset{a \in \mathcal{A}^{(k)}}\argmax\, Q(S_{t+1}^{(k)},a;\boldsymbol{\theta}_{t}^{(k)});\boldsymbol{{\grave{\theta}}}_{t}^{(k)})]^{2},
\end{aligned}
\label{eq:L_DDQN}
\end{equation}
We note that in (\ref{eq:L_DDQN}), the selection of action with the $\argmax$ function is done using the (main) DQN. 
This means that, as in Q-learning, the value of the policy is still estimated according to the parameter set $\boldsymbol{\theta}$. However, the second set of parameters $\boldsymbol{{\grave{\theta}}}$ is used to  fairly evaluate the value of this policy. As we observe from the numerical evaluation in Section~\ref{Results}, the introduction of the target network with the second parameter set reduces the overestimation of the Q-values, leading to better learning for the agent and, eventually, better performance in the context of V2X spectrum sharing.


\subsection{Transfer \textit{Q}-learning}\label{TQL}
 
In transfer Q-learning (\acrshort{TQL}), a previously trained \textit{expert model} is used to help a \textit{learner model} to learn the Q-values more efficiently by exploiting previously acquired knowledge. In particular, in TQL, the Q-values of the expert model will be transferred to the Q-values of the learner model. Therefore, the expert model can directly influence the Q-learning and convergence rate of the learner model. The main purpose of using such a strategy for the training of the agents is to provide the agents with prior information about the environment learned by the expert model beforehand, resulting in a decrease in training complexity. 

In the proposed framework, the expert model learns its policies using the \acrshort{DDQN} algorithm (\ref{eq:L_DDQN}). The expert model captures local environment information during its learning phase according to (\ref{eqn:Modified_State_Eq}) and performs action selection based on its own Q-values.  
The learner model uses the Q-values from the expert model to adjust its network parameters in the direction that improves system performance. We assume that the complete knowledge of the expert model is available to the learner model before the  start of the learning process. The use of the trained expert model improves the convergence of the DDQN algorithm and stabilizes the learning process for the learner model. 
Analogous to (\ref{eq:L_DQN}) and (\ref{eq:L_DDQN}), the Q-values from the expert model are used in the loss function to guide the learner model, resulting in the following formulation of the  
loss function

\begin{equation}
 \label{eqn:TQL_Loss_Eq}
\begin{aligned}
\mathcal{L}_{\mathrm{DDQN-TQL}}^{(k)}= & \sum _{\mathcal{B}^{(k)}}  [R_{t+1} -Q(S_{t}^{(k)},A_{t}^{(k)};\boldsymbol{\theta}_{t}^{(k)}) \\
     & +  \gamma \, Q(S_{t+1}^{(k)},\underset{a \in \mathcal{A}^{(k)}}\argmax\, Q(S_{t+1}^{(k)},a;\boldsymbol{\theta}_{t}^{(k)});\boldsymbol{{\grave{\theta}}}_{t}^{(k)})]^{2}\\
     & + Q_{exp}(S_{t}^{(k)},A_{t}^{(k)};\boldsymbol{\check{\theta}}_{t}^{(k)})]^{2},
\end{aligned}
\end{equation}
where, $Q_{exp}$ are the \textit{Q}-values from the expert model with parameter set $\boldsymbol{\check{\theta}}$.

\subsection{Evaluation in a realistic V2X environment}
\label{sec:ChannelModel}
As discussed in the Introduction, for the evaluation of the proposed framework in a realistic vehicular setting, we will resort to the 
$\mathrm{GEMV}^2$ model that intrinsically  supports  time  evolution  and  spatial consistency, while simultaneously adopting location-specific propagation
modeling with respect to large objects in the vicinity
of the communicating vehicles. 
$\mathrm{GEMV}^2$ uses 
simple geographical descriptors of the simulated environment
(outlines of buildings, foliage, and vehicles on the road) to
classify V2V links into three groups: (i) Line of sight (LOS)—links that have an unobstructed optical
path between the transmitting and receiving antennas; (ii) Non-LOS due to vehicles (NLOSv)—links whose LOS is
obstructed by other vehicles; (iii) Non-LOS due to buildings/foliage (NLOSb)—links
whose LOS is obstructed by buildings or foliage.
Furthermore, $\mathrm{GEMV}^2$ employs a
simple geometry-based small-scale signal variation model that
stochastically calculates the signal variation based on the information
about the surrounding objects.

$\mathrm{GEMV}^2$ can use vehicle
locations available from traffic mobility models (e.g., SUMO~\cite{IEEEexample:SUMO2018}) or real-world traces (e.g., via GPS) and the building and
foliage outlines and locations that are freely available from
projects such as OpenStreetMap~\cite{IEEEexample:OpenStreetMap}.


\section{Simulation Results}
\label{Results}

Subsequently, we present simulation results that demonstrate the advantages of the \acrshort{DDQN} and \acrshort{DDQN}-\acrshort{TQL} approaches presented in Section \ref{Contribution}. 

\begin{figure}
  \centering
  \includegraphics[width=7.5cm, height=7cm]{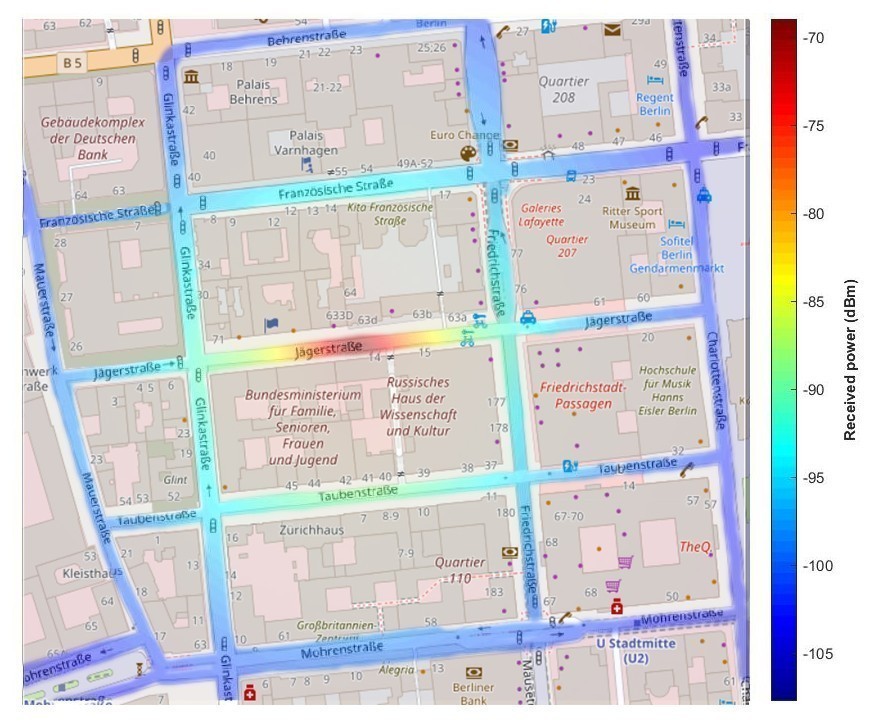}
    \caption{Simulation area in the center of Berlin. The colors indicate the receive powers with respect to an arbitrarily selected transmitter location. (© OpenStreetMap contributors) }
      \label{fig:Map} 
\end{figure}

As mentioned before, our goal is to model the \acrshort{V2X} environment as accurate as possible. Therefore, we use the $\mathrm{GEMV}^2$ model to obtain spatially consistent channels  (cf. Section \ref{sec:ChannelModel}). We use \acrshort{OSM} data from the center of Berlin, as shown in Fig.\ref{fig:Map} and we let  \acrshort{SUMO} generate random routes along the street grid for the specified simulation duration. The \acrshort{SUMO} traces for each vehicle are imported into the $\mathrm{GEMV}^2$ based MATLAB simulator, which combines them with  the geometric features from \acrshort{OSM} to generate both \acrshort{V2V} and \acrshort{V2I} channels. This is additionally illustrated in Fig.\ref{fig:Map}, where  different colors indicate the received power generated by $\mathrm{GEMV}^2$ at all locations along the street grid with respect to an arbitrarily selected transmitter location. 

The \textit{Q}-network for each agent is a five-layer fully connected neural network that comprises 3 hidden layers, with \acrfull{ReLU} as the activation function. We also use the $\epsilon$-greedy strategy to balance between exploration and exploitation, and we use the RMSProp optimizer \cite{IEEEexample:ruder2016overview} to update network parameters for training. We train each agent’s \textit{Q}-network in an episodic pattern. Each episode is spanning over the V2V payload delivery time $T$ and starts with a full V2V payload of size $B$. Thereby, the exploration rate $\epsilon$ is decreased linearly from $1$ to $0.02$ over the first $80\%$ of the episodes and remains constant afterward. 
During the training phase, we fix the large-scale fading for 100 training episodes and let the small-scale fading change over each time-step $t$. As in \cite{IEEEexample:liang2019spectrum}, we do this to support the agents in acquiring the underlying fading dynamics of the channel.

In the evaluation phase, however, the vehicles update their large-scale fading after every V2V payload delivery time $T$. Moreover, the values of $\epsilon$ and the iteration number $i$ are set to the values of the last step of the training phase. For a single simulation run, the traces and channels experienced by each vehicle are captured for 13 seconds, where the first 3 seconds are used for the training phase, and the remaining 10 seconds are used for the evaluation phase. The plots shown in this work are averaged over multiple simulation runs, where in each simulation run, the vehicles are assigned a new random \acrshort{SUMO} trace (with a different starting position). The detailed simulation parameters can be found in Table \ref{tab1}.

\begin{table}
\caption{Simulation parameters.}
\begin{center}
\begin{tabular}{|c|c|}
\hline
\textbf{Parameter}& \textbf{Value} \\
 \hline
Carrier frequency $f_{c}$& 2 GHz \\
 \hline
Bandwidth \textit{W} & 4 MHz \\
\hline
Number of \acrshort{V2I} links $M$ & 4 \\
\hline
Number of \acrshort{V2V} links $K$ & 4 \\
\hline
\acrshort{V2I} transmit power  $P_{m}^{c}$& 23 dBm \\
\hline
\acrshort{V2V} transmit power  $P_{k}^{d}$& [23,15,5,-100] dBm \\
\hline

V2V payload delivery time $T$ & 100 ms \\
\hline
Noise power $\sigma ^{2} $ & -114 dBm \\
\hline
\acrshort{V2V} payload size $B$ & [1,2,...] x 1060 bytes \\
\hline
Number of simulation runs & 5 \\
\hline

\end{tabular}\vspace{-.65cm}
\label{tab1}
\end{center}
\end{table}

\subsection{Performance of Double DQN} \label{sec:results_DDQN}
In the following, we compare \acrshort{MARL} based on \acrshort{DQN} (as proposed in \cite{IEEEexample:liang2019spectrum}) with the proposed \acrshort{DDQN} approach, introduced in Section \ref{DDQN}. In addition, we show the performance of the same random baseline as in \cite{IEEEexample:liang2019spectrum}, which selects transmission powers and spectrum sub-bands randomly. We train each agent’s \textit{Q}-network for a total of $3000$ episodes.

\begin{figure}

\begin{minipage}[t]{0.5\linewidth}
    \includegraphics[width=1\linewidth,height=1\textheight,keepaspectratio]{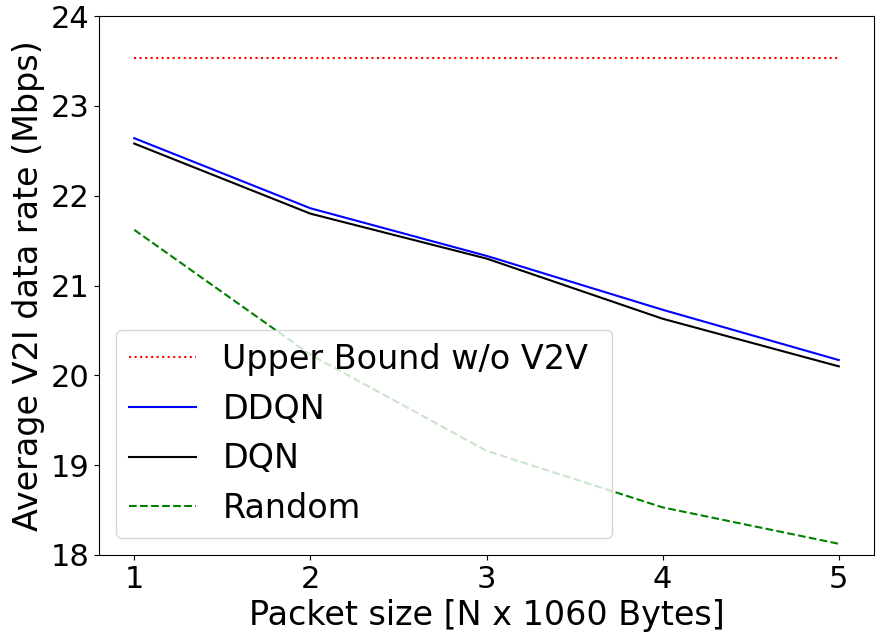}
    \subcaption{Sum capacity of \acrshort{V2I} links.}
    \label{fig:DDQN_Fig1}
\end{minipage}%
\begin{minipage}[t]{0.5\linewidth}
    \includegraphics[width=1\linewidth,height=0.3\textheight,keepaspectratio]{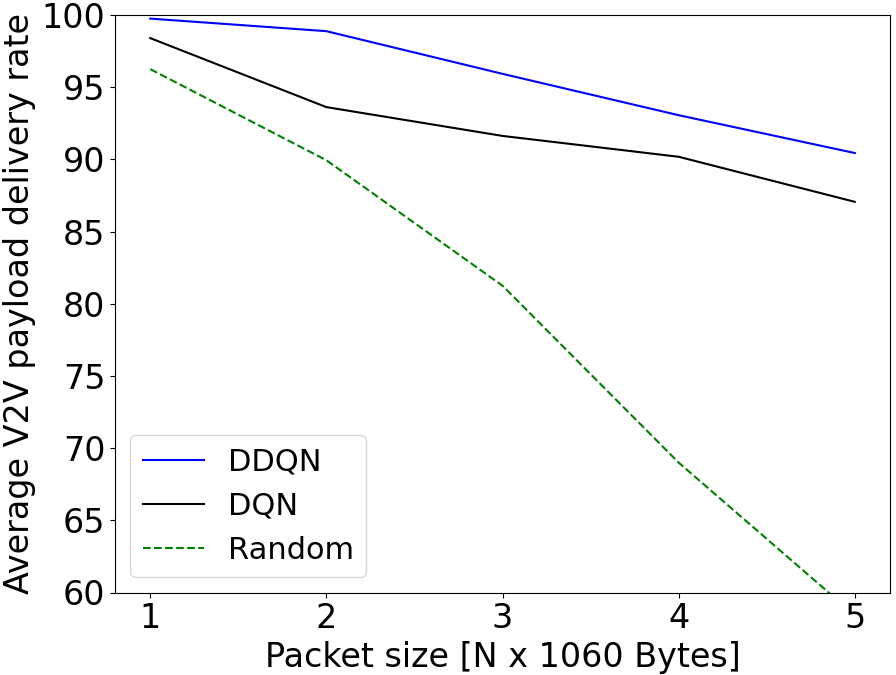}
    \subcaption{\acrshort{V2V} delivery rate.}
    \label{fig:DDQN_Fig2}
\end{minipage} 
\begin{minipage}[t]{0.49\linewidth}
    \includegraphics[width=1\linewidth,height=1\textheight,keepaspectratio]{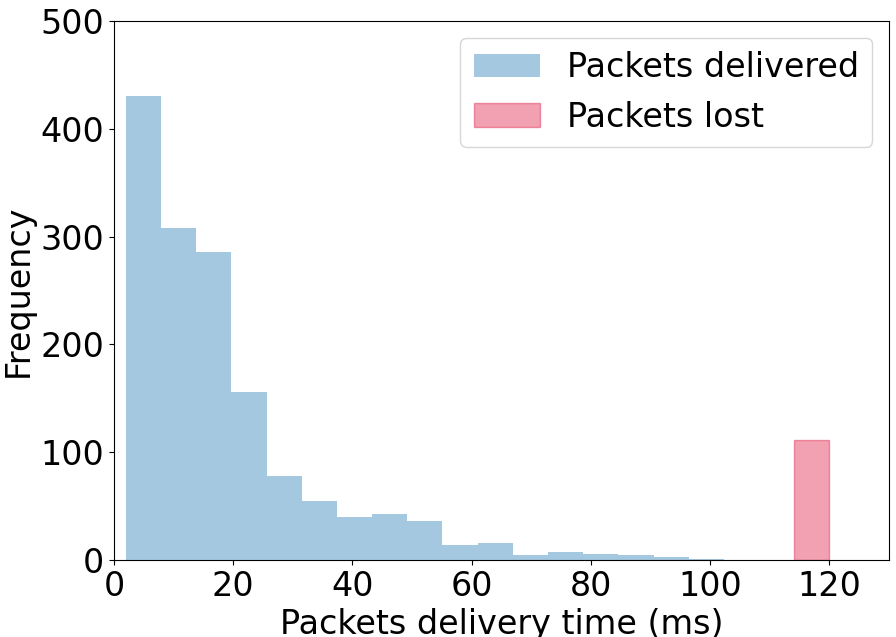}
    \subcaption{Histogram of \acrshort{V2V} packet delivery times for \acrshort{DDQN}.}
    \label{fig:Hist_DDQN}
\end{minipage} 
\begin{minipage}[t]{0.49\linewidth}
    \includegraphics[width=1\linewidth,height=1\textheight,keepaspectratio]{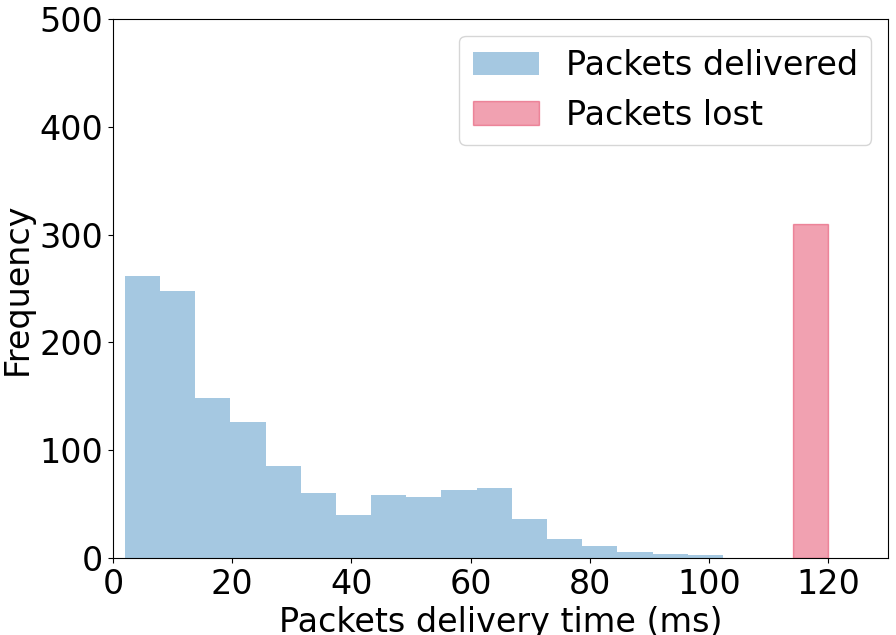}
    \subcaption{Histogram of \acrshort{V2V} packet delivery times for \acrshort{DQN}.}
    \label{fig:Hist_DQN}
\end{minipage} 

\caption{Comparison of DDQN, DQN and random baseline over different payload sizes.}
\end{figure}

Fig.\ref{fig:DDQN_Fig1} shows the \acrshort{V2I} performance with increasing payload size $B$. At the end of the evaluation phases, average data rates of \acrshort{V2I} links (shown in Fig.\ref{fig:DDQN_Fig1}) and the \acrshort{V2V} payload delivery rate (shown in Fig.\ref{fig:DDQN_Fig2}) are computed. The increase of the \acrshort{V2V} payload not only results in a longer \acrshort{V2V} transmission duration, as it becomes more challenging to completely deliver the payload within the delivery time limit, but it also leads to an increased interference to \acrshort{V2I} links. This in turn results in a reduced \acrshort{V2I} sum capacity, as we can observe in Fig.\ref{fig:DDQN_Fig1}. 

 While the \acrshort{V2I} performance for the \acrshort{DDQN} approach is close to the \acrshort{DQN} approach, we observe in Fig.\ref{fig:DDQN_Fig2}  that the proposed approach improves the success probability compared to the \acrshort{DQN} approach by up to $4\%$. As expected, the success probability drops for higher payload sizes, where the random baseline method shows a much more severe degradation, compared to the reinforcement learning-based  algorithms.

The design of the reward function encourages the agents to finish their \acrshort{V2V} packet transmission in a timely manner. Fig.\ref{fig:Hist_DDQN} and Fig.\ref{fig:Hist_DQN} show histograms of the observed packet delivery times for the proposed \acrshort{DDQN}-based algorithm and the \acrshort{DQN}-based baseline from \cite{IEEEexample:liang2019spectrum}, respectively. Here, we fix the payload size to $4\times1060$ bytes, and all packets delivered after the packet delivery time limit (i.e, $T=100$\,ms) are considered to be lost (accumulated in the red bars). Clearly, the \acrshort{DDQN} approach leads to earlier completion of transmissions, which results in a significantly reduced number of failed packet deliveries. For the given simulation settings, the median packet delivery time for successful V2V transmission using \acrshort{DDQN} is $14$\,ms as compared to $19$\,ms for  \acrshort{DQN}. 
In summary, the results indicate that the 
reduction in overestimation of \textit{Q}-values due to the \acrshort{DDQN} approach positively influences the learning for the agents. 
\subsection{Performance of Transfer Q-Learning}
Subsequently, we evaluate the learning performance with, and without the prior knowledge of an expert model (as explained in Section \ref{TQL}). For this, we compare the same \acrshort{DDQN}-based method, as evaluated in the previous subsection, with the proposed  \acrshort{DDQN}-\acrshort{TQL} method.  The state space, action space and reward function for both the expert model and the learner model are identical. 
To evaluate the adaptivity of \acrshort{DDQN}-\acrshort{TQL}, we use different channel conditions for the training of both models.
The main motivation in using \acrshort{DDQN}-\acrshort{TQL} is to reduce the required training duration. To evaluate this, we reduce the training duration of the learner model to 1,800 episodes in our performance evaluations (while we keep the training duration of the expert model at 3,000 episodes).

\begin{figure}

\begin{minipage}[t]{0.5\linewidth}
    \includegraphics[width=1\linewidth,height=1\textheight,keepaspectratio]{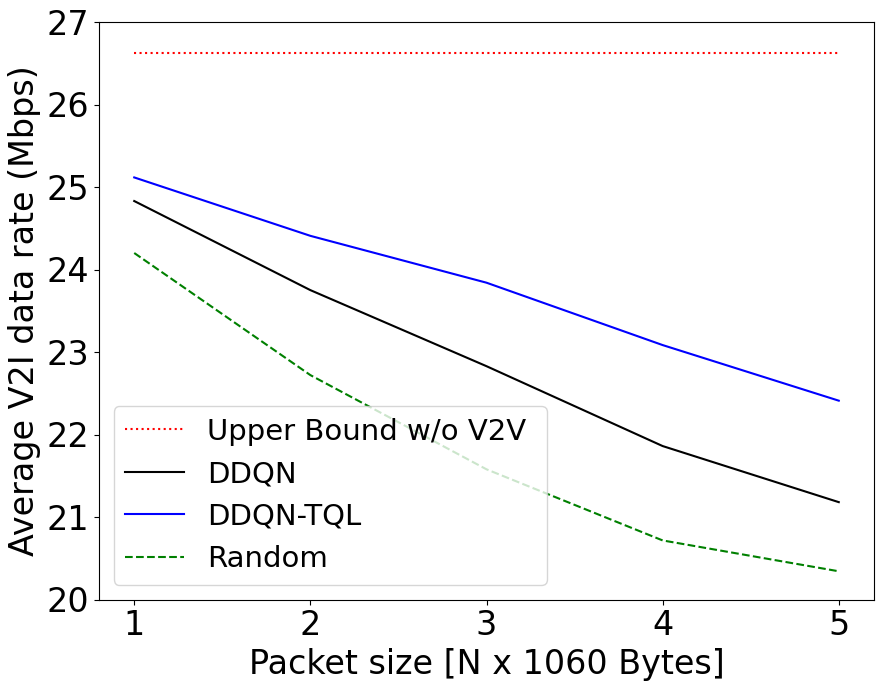}
    \subcaption{Sum capacity of \acrshort{V2I} links.}
    \label{fig:TQL_Fig1}
\end{minipage}%
\begin{minipage}[t]{0.5\linewidth}
    \includegraphics[width=1\linewidth,height=1\textheight,keepaspectratio]{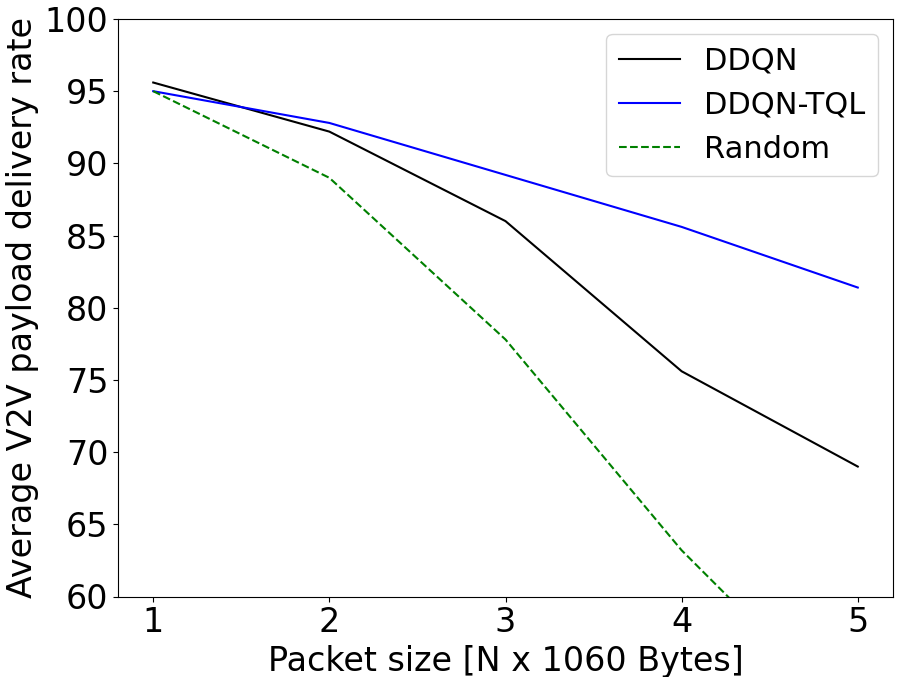}
    \subcaption{\acrshort{V2V} delivery rate.}
    \label{fig:TQL_Fig2}
\end{minipage} 
\begin{minipage}[t]{0.49\linewidth}
    \includegraphics[width=1\linewidth,height=1\textheight,keepaspectratio]{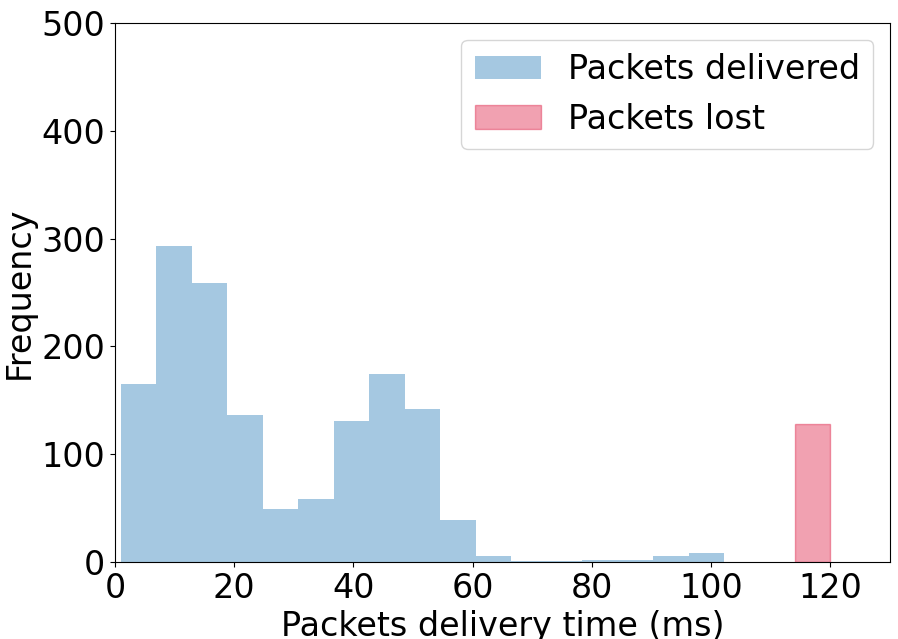}
    \subcaption{Histogram of \acrshort{V2V} packet delivery times for \acrshort{DDQN}-\acrshort{TQL}.}
    \label{fig:Hist_TQL}
\end{minipage} 
\begin{minipage}[t]{0.49\linewidth}
    \includegraphics[width=1\linewidth,height=1\textheight,keepaspectratio]{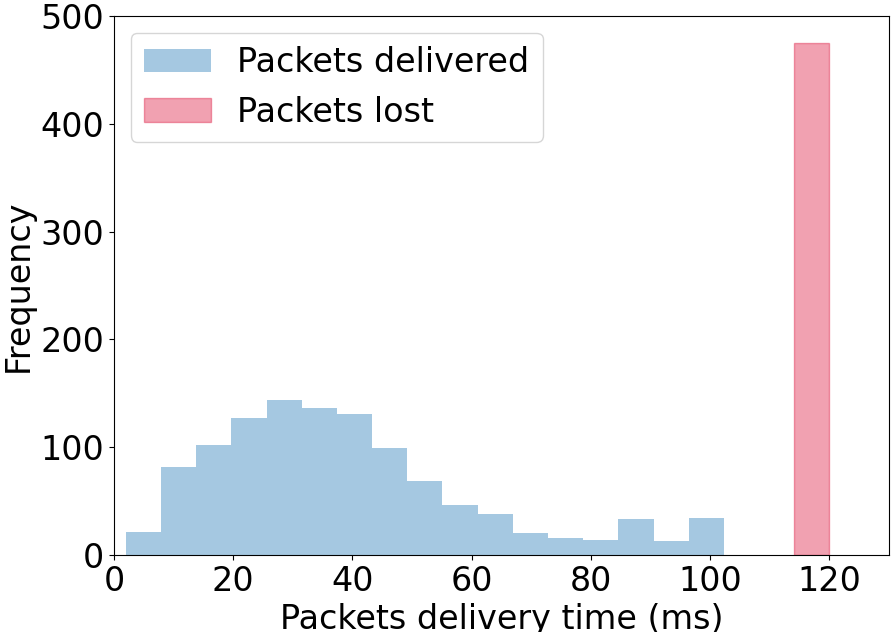}
    \subcaption{Histogram of \acrshort{V2V} packet delivery times for \acrshort{DDQN}.}
    \label{fig:Hist_TQLDDQN}
\end{minipage} 
\caption{Comparison of DDQN-TQL, DDQN and random baseline over different payload sizes.}
\label{fig:TQL}
\end{figure}

The experimental results in Fig.\ref{fig:TQL_Fig1}, which  shows the average sum capacity of \acrshort{V2I} links, support this reasoning. We can observe that  \acrshort{DDQN}-\acrshort{TQL} outperforms \acrshort{DDQN} for all considered payload sizes. While  for small payload sizes the performance differences are small, for high payload sizes,  \acrshort{DDQN}-\acrshort{TQL} provides significant gains in terms of \acrshort{V2I} data rates. In Fig.\ref{fig:TQL_Fig2}, which shows the   \acrshort{V2V} payload delivery rate, we observe a similar behavior. Especially for large packet sizes, \acrshort{DDQN}-\acrshort{TQL} enables the agents to learn  policies within the given training time that show significantly higher performance than the policies obtained without the \textit{Q}-value transfer. For example, for a payload size of $4\times1060$ bytes, we observe performance gains of $6\%$ and $12\%$,   for the \acrshort{V2I} sum capacity, and the \acrshort{V2V} payload delivery rate, respectively. 

The packet delivery times for  \acrshort{DDQN}-\acrshort{TQL} and \acrshort{DDQN} are provided in greater detail in Fig.\ref{fig:Hist_TQL} and Fig.\ref{fig:Hist_TQLDDQN}, respectively. Again, we fix the payload size to $4\times1060$ bytes. The changes in the empirical distribution of the packet delivery times demonstrate the effectiveness of the knowledge transfer approach for the given training duration.  The \acrshort{TQL}-based approach leads to  policies that finish transmissions earlier and thus reduce the number of failed packet deliveries. More precisely, for the simulations shown in Fig. \ref{fig:TQL}, the median packet delivery time for successful V2V transmission using \acrshort{DDQN}-\acrshort{TQL} is $18$\,ms, compared to $35$\,ms for \acrshort{DDQN}. 

We would like to point out that the performance of the proposed method highly depends on the choice of the number of training episodes. For a very short training duration, agents are unable to properly explore the environment, and reinforcement learning-based  schemes will not even outperform the random baseline. On the other hand, learning for very long training intervals can lead to a potential  over-fitting, which is a common problem for deep \acrshort{RL} algorithms  \cite{IEEEexample:hardt2016train}\cite{IEEEexample:zhang2018study}. As the \acrshort{TQL} approach is more effected by over-fitting, 
simulations show that for larger numbers of  training episodes it is outperformed by  \acrshort{DDQN}, where we train our model from scratch without any previous knowledge about input-output relations. 

\begin{figure}

\begin{minipage}[t]{0.5\linewidth}
    \includegraphics[width=1\linewidth,height=1\textheight,keepaspectratio]{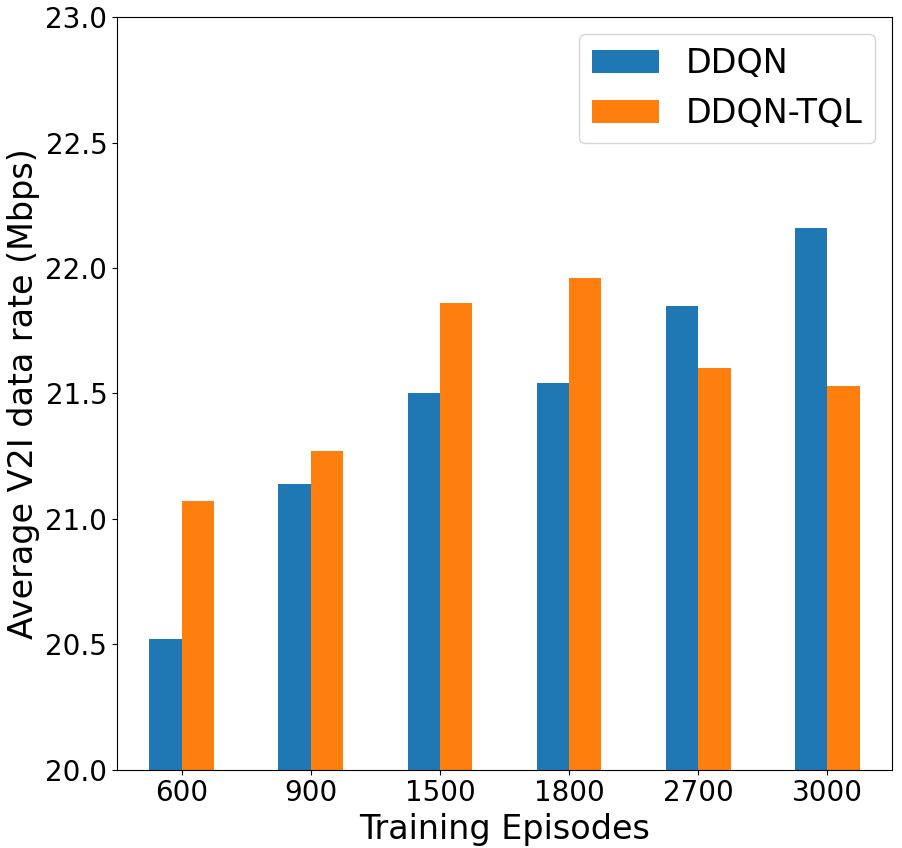}
    \subcaption{Sum capacity of \acrshort{V2I} links.}
    \label{fig:Bar1}
\end{minipage}%
\begin{minipage}[t]{0.5\linewidth}
    \includegraphics[width=1\linewidth,height=1\textheight,keepaspectratio]{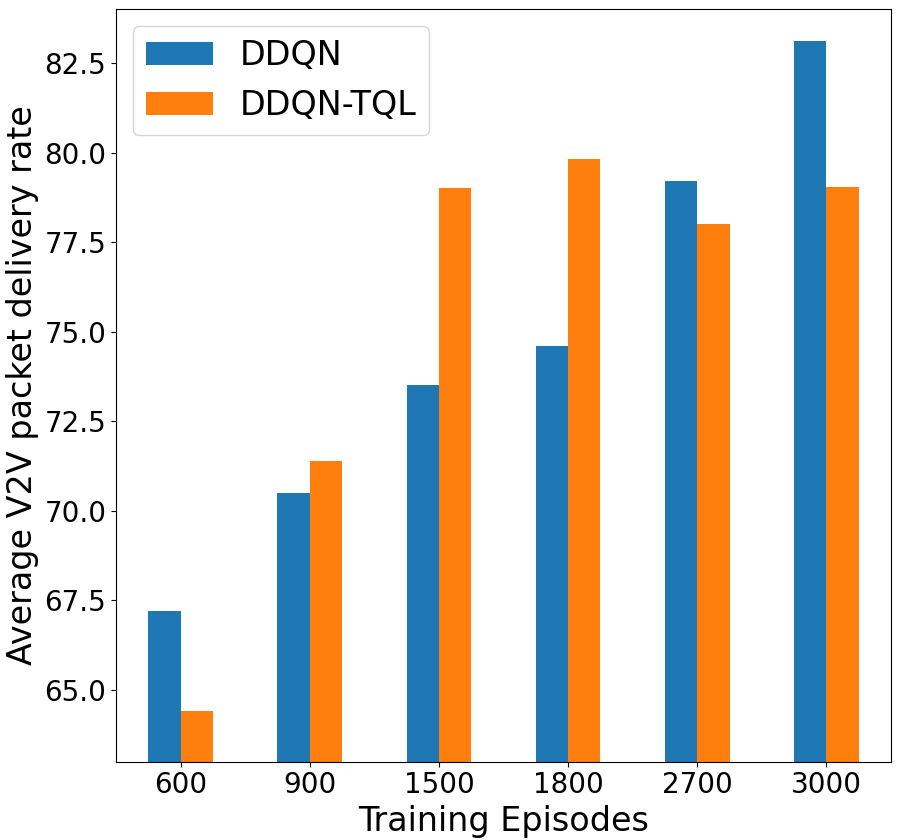}
    \subcaption{\acrshort{V2V} delivery rate.}
    \label{fig:Bar2}
\end{minipage} 

\caption{ Comparison of DDQN-TQL and DDQN for different numbers of training episodes.}
\label{fig:barplot}
\end{figure}

This behaviour is illustrated in Fig.\ref{fig:barplot}. The performance with respect to \acrshort{V2I} (Fig.\ref{fig:Bar1}) and \acrshort{V2V} (Fig.\ref{fig:Bar2}) links are shown for a fixed payload size of $4\times1060$ bytes. The figure confirms the previous discussions, and we observe that especially in scenarios with only a reduced training time around $1800$ training episodes, the \acrshort{TQL} approach provides significant advantages. 

\section{Conclusion}
\label{Conclusion}
In this paper, we proposed improvements and extensions to deep Q-learning-based approaches (such as the \acrshort{MARL} algorithm in \cite{IEEEexample:liang2019spectrum}) to distributed spectrum sharing in \acrshort{V2X} communications. In particular, we proposed and evaluated solutions for reducing the overestimation of Q-values (using the Double DQN method), and for reducing the training time (using a transfer Q-learning approach). In contrast to previous studies, we evaluated the proposed methods using a realistic vehicular setting based on the  geometry-based deterministic $\mathrm{GEMV}^2$ channel model.   
Simulations show that the use of   \acrshort{DDQN} indeed results in improved learning performance of the agents. Moreover, we observe that the performance of the proposed \acrshort{DDQN} method with \acrshort{TQL} is significantly less susceptible to a shorter training duration than without \acrshort{TQL}. 

\bibliographystyle{./IEEEtran}

\bibliography{./ms}

\end{document}